**Cloud-top infrared observations reveal the four-dimensional precipitation structure**

Tianchi Xu[1], Ziqiang Ma[1*], Andrea Marinoni[2], Yuanpeng He[3], Xiaoqing Li[4], Chuanfeng Zhao[5], Kang He[1], Jintao Xu[5], Bohan Zhou[3], Wenbo Zhao[1], Haoshuang Chen[1], Tun Wang[6], Dongdong Wang[1], Yang Hong[7]

1 Institute of Remote Sensing and Geographical Information Systems, School of Earth and Space Sciences, Peking University, Beijing, 100871, China

2 Department of Computer Science and Technology, University of Cambridge, Cambridge, United Kingdom

3 Key Laboratory of High Confidence Software Technologies (Peking University), Ministry of Education; School of Computer Science, Peking University, Beijing, 100871, China

4 Innovation Center for Fengyun Meteorological Satellite, National Meteorological Centre, China Meteorological Administration, Beijing, 100081, China

5 Department of Atmospheric and Oceanic Sciences, School of Physics, Peking University, Beijing, 100871, China

6 Institute of Mountain Hazards and Environment, Chinese Academy of Sciences, Chengdu, 610213, China

7 School of Civil Engineering and Environmental Science, University of Oklahoma, Norman, OK, United States

* Correspondence authors: Dr. Ziqiang Ma (ziqma@pku.edu.cn);

## Highlight

1. Cloud-top infrared observations encode information on sub-cloud precipitation structure.

2. Four-dimensional precipitation evolution can be reconstructed from geostationary infrared measurements.

3. This challenges the conventional view that infrared observations are intrinsically insensitive to precipitation beneath clouds.

**Abstract**

Accurate four-dimensional (4D) precipitation information is essential for understanding the Earth's energy and water cycles, yet remains observationally unresolved at global scales. Conventional theory holds that geostationary infrared observations primarily sense cloud-top properties, with limited sensitivity to sub-cloud precipitation. Here we show that cloud-top infrared measurements nevertheless encode sufficient information to recover the four-dimensional structure of precipitation, revealing a previously unexploited observability of sub-cloud processes. We introduce a physically constrained deep learning framework, 4DPrecipNet, in which a moisture-first constraint requires the latent representation to recover precipitable water vapour, anchoring the model in thermodynamic consistency. By integrating multi-channel infrared radiances with these constraints and radar-derived precipitation profiles, we reconstruct the vertical and temporal evolution of precipitation systems from geostationary orbit. The framework captures deep convective structures and their evolution, with robust performance across large samples and independent radar comparisons. These results demonstrate that sub-cloud precipitation is physically encoded in cloud-top infrared observations, establishing a new pathway for continuous global monitoring of precipitation structure.

## 1. Introduction

Accurate characterization of the four-dimensional (4D) structure of precipitation is fundamental to the Earth's energy and water cycles, yet remains poorly constrained at global scales (Loeb et al., 2024; Bodnar et al., 2025; Ma et al., 2025; Su et al., 2026). Although geostationary infrared observations provide continuous, high-frequency coverage of the atmosphere, they are widely assumed to sense only cloud-top properties and to be intrinsically insensitive to precipitation beneath clouds (Sorooshian et al., 2000; Lu and Wu, 2000; Ma et al., 2022; Zhu and Ma, 2022; He et al., 2025). This limitation has long constrained their use for resolving the vertical structure of precipitation systems. Here we show that cloud-top infrared observations encode sufficient information to reconstruct the four-dimensional structure of precipitation, challenging this conventional view.

Existing approaches to observing precipitation structure rely primarily on active radar measurements and passive microwave retrievals (Hou et al., 2014; Pfreundschuh et al., 2024; Xu et al., 2025). Spaceborne precipitation radars provide direct vertical profiling but suffer from sparse spatial coverage and limited temporal sampling, preventing continuous monitoring of rapidly evolving systems (Hou et al., 2014; Skofronick-Jackson et al., 2017; Seto et al., 2021; Ali et al., 2026). Passive microwave observations offer broader coverage but depend on indirect radiative signatures that are strongly affected by surface emissivity and microphysical assumptions, introducing substantial uncertainties, particularly over land (Boukabara et al., 2011; Xu et al., 2025; Ryu et al., 2025) . In contrast, geostationary infrared observations provide continuous, high-frequency global coverage (Yang et al., 2017; Yang et al., 2023; Schmit et al., 2023), yet have been largely confined to cloud-top diagnostics and empirical precipitation estimation, without resolving the vertical structure of precipitation (Hong et al., 2004; Lu et al., 2024; He et al., 2025; Nosratpour et al., 2025).

Consequently, a fundamental gap remains between temporally continuous geostationary observations and vertically resolved precipitation information.

Here we introduce a physically constrained framework, 4DPrecipNet, that enables recovery of the four-dimensional structure of precipitation from geostationary infrared observations. Central to the framework is a moisture-first constraint, in which the latent representation is required to recover precipitable water vapor (PWV) under ERA5 supervision (Hersbach et al., 2020), thereby anchoring the network in Clausius–Clapeyron thermodynamics (Da Silva and Haerter, 2025). As PWV represents the vertically integrated moisture reservoir that fundamentally constrains precipitation formation, its reconstruction provides a physically meaningful constraint on the latent thermodynamic state. This formulation converts the learning process from an unconstrained statistical mapping into a thermodynamically guided inference problem. By integrating multi-channel infrared radiances with physically consistent atmospheric constraints, the framework establishes a direct and interpretable linkage between cloud-top radiative signatures and sub-cloud precipitation processes, revealing information that has remained inaccessible in conventional infrared analyses. Unlike empirical infrared-based precipitation estimation, this approach exploits the intrinsic physical information embedded in infrared observations to resolve precipitation structure (Reichstein et al., 2019; Karniadakis et al., 2021). Applied at geostationary scales, it captures the vertical organization and temporal evolution of precipitation systems, enabling continuous, high-frequency four-dimensional monitoring while fundamentally alleviating the conventional "black-box" critique of deep learning (Reichstein et al., 2019).

We evaluate this framework, 4DPrecipNet, using extensive co-located satellite and radar observations spanning diverse precipitation regimes. The results show that it reproduces the vertical organization and temporal evolution of precipitation systems, including deep convective structures and their development. Comparisons with independent spaceborne radar observations demonstrate consistent agreement in the retrieved vertical profiles, while large-sample analyses indicate robust performance across a wide range of conditions. These results provide direct empirical evidence that cloud-top infrared observations can resolve the four-dimensional structure of precipitation, bridging the long-standing gap between temporally continuous geostationary observations and vertically resolved precipitation information.

## 2. Observability of precipitation structure from cloud-top infrared observations

Conventional radiative-transfer reasoning suggests that geostationary infrared observations cannot robustly constrain the three-dimensional structure of hydrometeors beneath the cloud-top, as cloud-top radiances encode precipitation only indirectly. A key unresolved question is therefore whether multi-channel infrared observations contain sufficient information to overcome this structural degeneracy. To establish the physical basis of the framework, we perform a channel-elimination experiment that serves as a linchpin of the observability argument (Fig. 1b). The results reveal a sharp transition in retrievability: when fewer than three spectral channels are retained, retrieval uncertainty increases rapidly and predicted distributions become diffuse, indicating ambiguity among competing precipitation structures. This explains why previous single- or dual-channel infrared approaches failed to constrain sub-cloud precipitation. In contrast, the joint use of multi-channel infrared observations collapses the solution space, substantially reducing uncertainty and enabling dynamically informative constraints to emerge. This behaviour indicates the existence of a spectral threshold beyond which infrared observations become capable of constraining precipitation structure. Complementary ablation experiments further demonstrate that the optimal thermodynamic constraint ($\lambda_{PWV}$ = 5) reduces positive bias by ~20%, highlighting the physical consistency of the framework. Guided by this result, we reformulate the problem as a physically constrained inference that links cloud-top radiative signals to sub-cloud precipitation processes (Raissi et al., 2019; Karniadakis et al., 2021) (Fig. 1a,c). Specifically, the reconstruction is decomposed into two coupled components: thermodynamic background inference and kinematics-driven vertical precipitation reconstruction. The thermodynamic component establishes a physically consistent moisture field that anchors the solution space, while the kinematic component maps this constrained state to vertically resolved precipitation structures.

This formulation departs from conventional end-to-end empirical mappings by explicitly embedding physical constraints into representation learning, enabling infrared observations to recover the four-dimensional structure of precipitation from geostationary orbit (Reichstein et al., 2019).

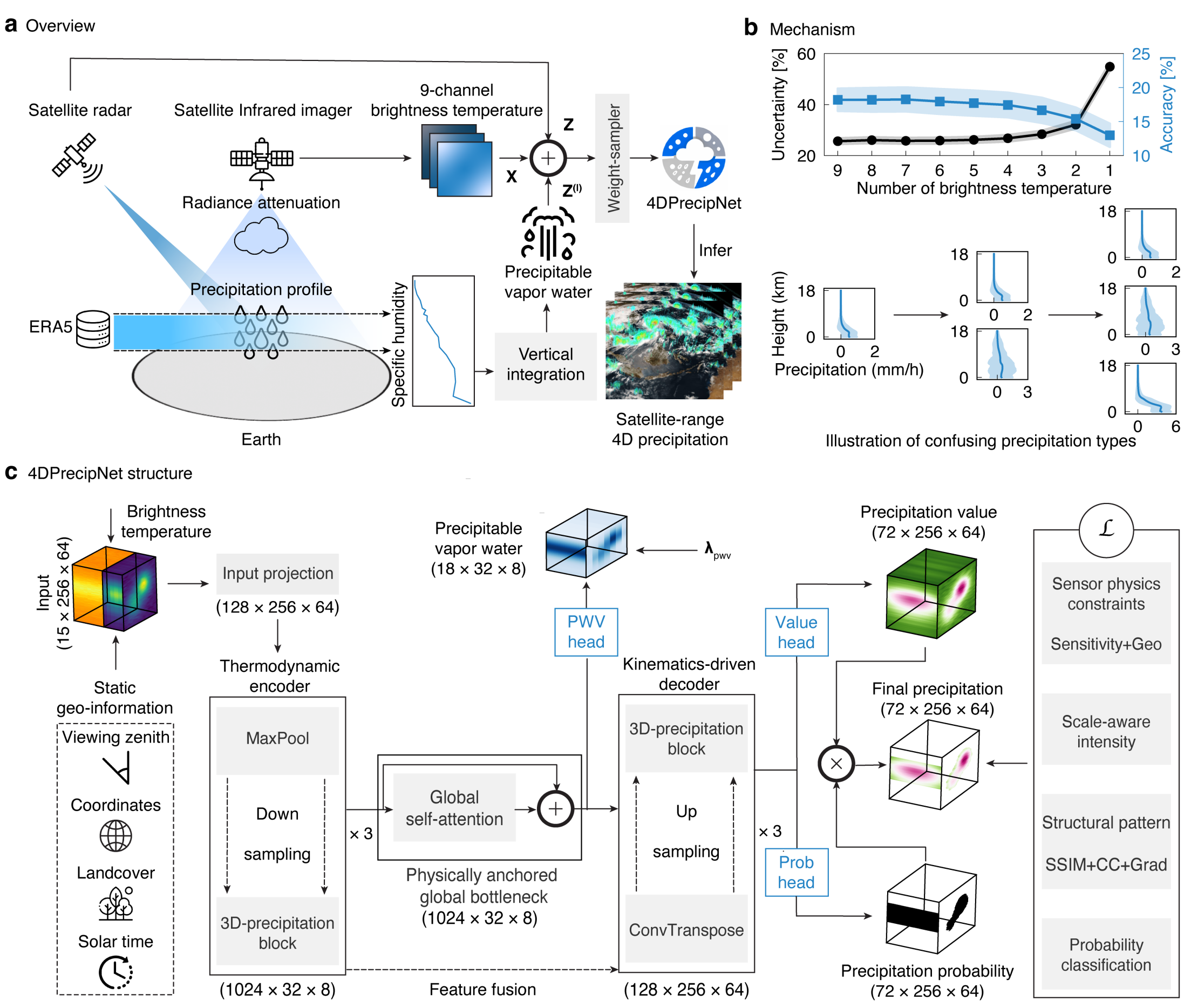

**Figure 1 | Infrared observability and physically constrained reconstruction of four-dimensional precipitation structure.**

**a,** Conventional view in which geostationary infrared observations are assumed to primarily sense cloud-top properties, providing limited information about precipitation beneath clouds.

**b,** Channel-elimination experiment revealing a transition in infrared observability. Retrieval uncertainty increases sharply when fewer than three spectral channels are used, indicating structural ambiguity, whereas the full set of channels collapses the solution space and reduces uncertainty, demonstrating that multi-channel infrared observations contain dynamically informative constraints on precipitation structure.

**c,** Physically constrained reconstruction framework, 4DPrecipNet, linking cloud-top radiative signals to sub-cloud precipitation processes. The formulation decomposes the problem into thermodynamic background inference, which establishes a physically consistent moisture field, and kinematics-driven reconstruction, which maps this constrained state to vertically resolved precipitation structure.

## 3. Resolving the three-dimensional precipitation structure in extreme tropical cyclones

To assess performance under extreme dynamical conditions, we analyze Super Typhoon Matmo (2025) as a representative case (Prakash and Mohapatra, 2024; Gupta and Arthur, 2025) (Fig. 2). 4DPrecipNet reconstructs a compact eyewall precipitation ring exceeding 40 mm $h^{-1}$ while effectively suppressing spurious signals from cold, non-precipitating cirrus that typically confound infrared-based retrievals (Hong et al., 2004) (Fig. 2a). This demonstrates that cloud-top infrared observations can discriminate dynamically active precipitation from radiatively similar but non-precipitating cloud structures.

Along transect A–B, the retrieved vertical structure closely matches collocated FY-3G precipitation radar observations (Zhang et al., 2019) (Fig. 2b). A deep convective tower centered near 114.08°E extends from the surface to above 12 km, capturing the full vertical development of intense convection (Houze, 2014). Notably, a sharp intensity transition around 4–5 km is consistent with a melting-layer signature, indicating that the framework implicitly resolves microphysical phase transitions from infrared observations alone (Houze, 2014).

Quantitative evaluation confirms that the reconstruction preserves both intensity and structure. The peak precipitation (29.0 mm $h^{-1}$) agrees closely with the radar reference (28.5 mm $h^{-1}$), and the mean absolute error along the transect is 1.28 mm $h^{-1}$ (Fig. 2c). Unlike conventional learning-based retrievals that exhibit spatial smoothing, the model retains sharp gradients and convective extremes (Ravuri et al., 2021), demonstrating its ability to recover physically meaningful vertical organization. Consistent behavior is observed for the structurally asymmetric Typhoon Ragasa (Supplementary Fig. 2), suggesting robustness across diverse storm morphologies.

Together, these results provide direct evidence that geostationary infrared observations can resolve the three-dimensional structure and evolution of precipitation, bridging the long-standing gap between continuous infrared monitoring and vertically resolved precipitation measurements.

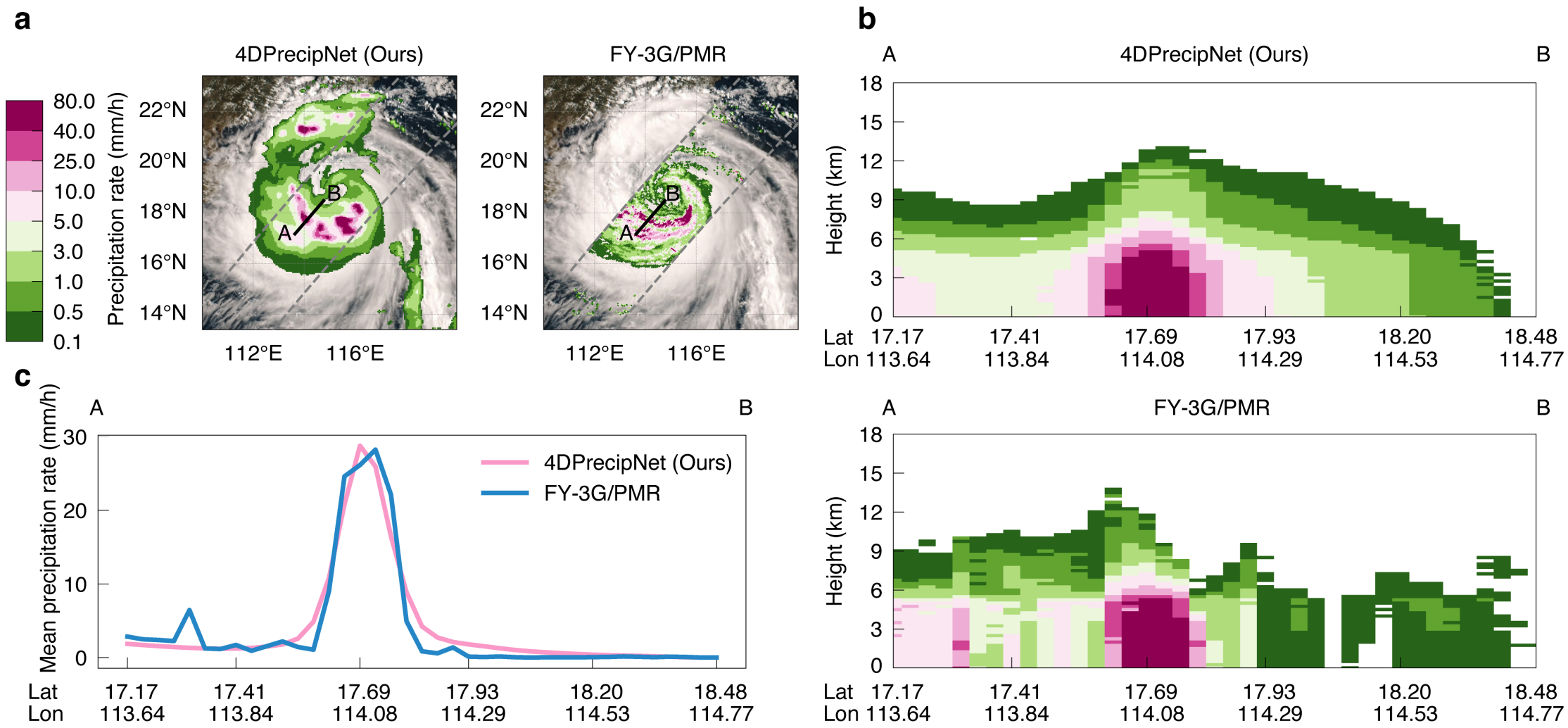

**Figure 2 | Retrieval of three-dimensional precipitation structure in an extreme tropical cyclone**

**a,** Spatial distribution of surface precipitation rate for Super Typhoon Matmo at 07:03 UTC on 4 October 2025. The 4DPrecipNet retrieval from geostationary multi-channel infrared observations (left) reproduces the compact eyewall and spiral rainband structure with strong spatial consistency relative to FY-3G precipitation radar measurements (right), while suppressing spurious signals from cold, non-precipitating cirrus.

**b,** Vertical precipitation structure along transect A–B. The infrared-based retrieval (top) captures a deep convective tower centered near 114.08°E, with precipitation extending from the surface to above 12 km, in close agreement with the radar reference (bottom). A pronounced intensity transition at ~4–5 km indicates a melting-layer signature, suggesting that the framework resolves microphysical phase transitions from cloud-top infrared observations.

**c,** Quantitative evaluation of mean precipitation intensity along the vertical cross-section. The retrieved peak intensity (29.0 mm $h^{-1}$) closely matches the radar-derived reference (28.5 mm $h^{-1}$), with a mean absolute error of 1.28 mm $h^{-1}$. Unlike conventional retrievals that exhibit spatial smoothing, the model preserves sharp gradients and convective extremes, demonstrating its ability to recover physically consistent vertical precipitation structure.

Together, these results provide direct observational evidence that cloud-top infrared measurements can resolve the three-dimensional structure of precipitation under extreme dynamical conditions.

## 4. Evaluation of three-dimensional precipitation structure within the geostationary satellite field of view

To evaluate structural skill under diverse meteorological conditions, we conducted an independent validation using more than 7.37 million vertical precipitation profiles from GPM Core Observatory Dual-frequency Precipitation Radar (DPR) overpasses between March 2024 and August 2025 (Hou et al., 2014; Skofronick-Jackson et al., 2017; Seto et al., 2021; Fig. 3). Profiles were stratified into four physically distinct regimes—deep convection, shallow convection, stratiform precipitation and weak precipitation—based on vertical structural characteristics, enabling regime-aware assessment of the model's ability to resolve precipitation structure (Houze, 2014; Aoki et al., 2026).

Model performance increases with precipitation intensity, with highest skill achieved for deep convection (CSI = 0.64, POD = 0.69, FAR = 0.06, CC = 0.57; Fig. 3a). Although absolute errors are larger in intense precipitation (MAE ≈ 6.71 mm $h^{-1}$), the sustained correlation indicates that the model preserves the spatial organization of convective systems rather than introducing excessive smoothing. Skill decreases toward weaker regimes, particularly for stratiform precipitation, highlighting the intrinsic challenge of light-precipitation detection from infrared observations. This behaviour can be explained physically: unlike deep convection, weak and stratiform systems are characterized by limited vertical kinematic forcing and relatively uniform cloud-top thermal structure, which reduces the sensitivity of infrared radiances to sub-cloud precipitation processes. In such regimes, cloud-top thermal gradients provide only weak constraints on sub-cloud precipitation, limiting the observability of precipitation signals from infrared measurements. The presence of extensive non-precipitating cirrus further obscures the signal, making it difficult to distinguish precipitating from non-precipitating clouds based solely on cloud-top infrared signatures. Despite these limitations, the model retains coherent structural signals across all regimes, indicating that physically constrained learning still extracts meaningful precipitation-related information even under weak forcing. At the full-disk scale, spatial evaluation against IMERG-Final V07B surface precipitation rate demonstrates consistent performance across the tropics and subtropics (30°S–30°N) (Huffman et al., 2020), with limited sensitivity to viewing geometry and no pronounced seasonal degradation (Supplementary Fig.

4). IMERG-Final represents a state-of-the-art multi-satellite precipitation product that benefits from passive microwave observations and retrospective gauge-based bias correction (Huffman et al., 2020), providing one of the most accurate and heavily corrected global precipitation benchmarks currently available. Compared with existing infrared-based precipitation products, 4DPrecipNet achieves systematically higher skill (Hong et al., 2004; Zhu et al., 2025), with domain-averaged POD reaching ~0.58—nearly doubling that of operational retrievals—while maintaining comparable or improved correlation (CC > 0.5). Notably, the fact that our infrared-only framework achieves such detection performance against this heavily corrected benchmark constitutes a strong testament to its capability to recover surface precipitation signals without direct access to microwave or gauge information. Together, these results demonstrate that the physically constrained framework, 4DPrecipNet, generalizes across diverse meteorological regimes and enables robust recovery of the three-dimensional structure of precipitation at global scales.

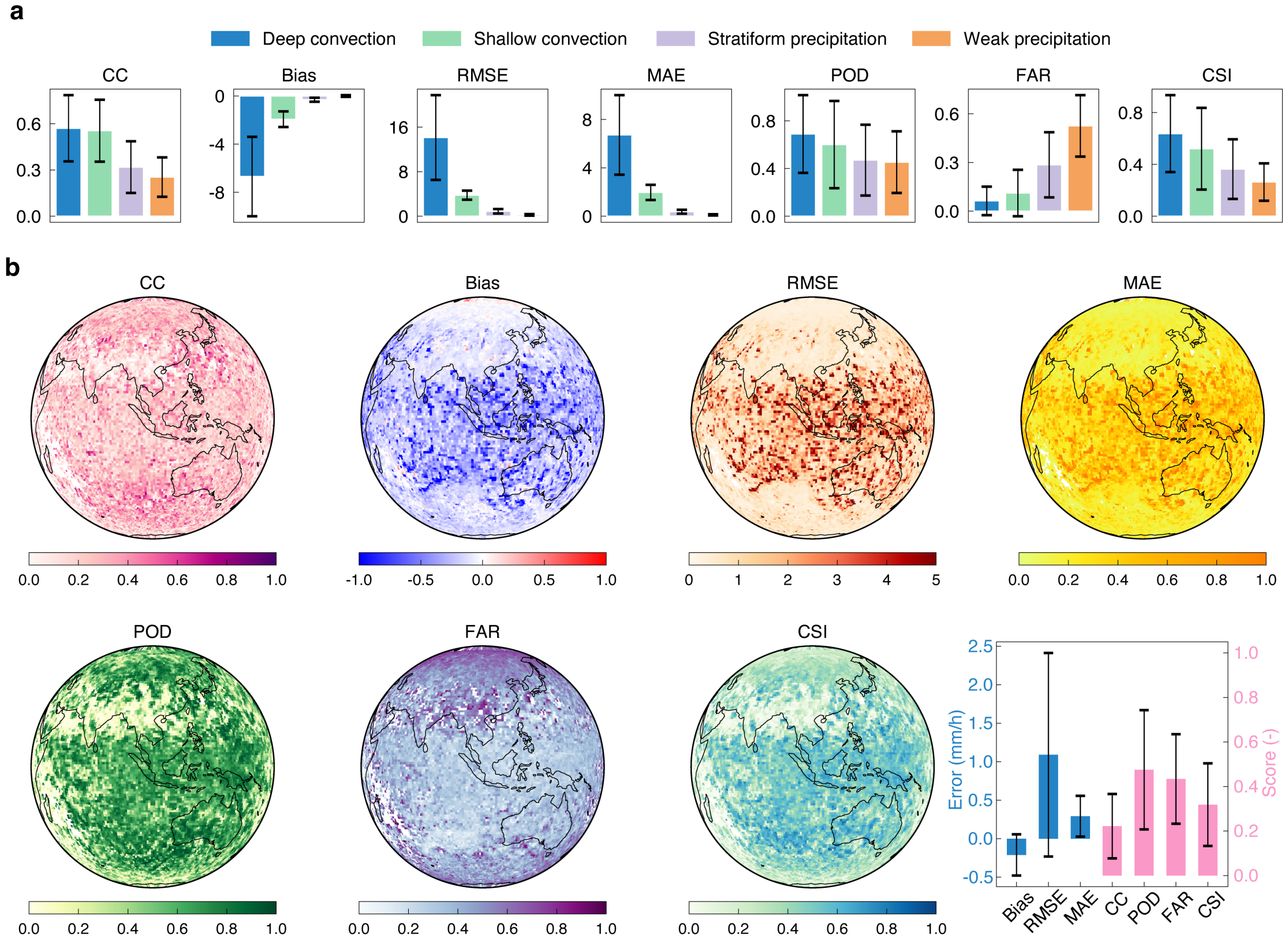

**Figure 3 | Global evaluation of three-dimensional precipitation structure reconstruction.**

**a,** Regime-stratified performance across 7.37 million vertical precipitation profiles from GPM Core Observatory Dual-frequency Precipitation Radar (DPR). Profiles are grouped into deep convection, shallow convection, stratiform precipitation and weak precipitation based on vertical structure. Model skill increases with precipitation intensity, with highest performance in deep convection, while retaining coherent structural signals across all regimes.

**b,** Spatial distribution of evaluation metrics at the full-disk (FD) scale of the geostationary satellite FY-4B, including correlation coefficient (CC), bias, RMSE, mean absolute error (MAE), probability of detection (POD), false alarm ratio (FAR) and critical success index (CSI). Results show spatially consistent skill in reconstructing precipitation structure across the tropics and subtropics, with limited sensitivity to viewing geometry.

**c,** Comparison with existing infrared-based precipitation products. 4DPrecipNet achieves systematically higher detection skill and comparable or improved correlation, while preserving structural coherence of precipitation fields.

## 5. Continuous monitoring of precipitation-system evolution from geostationary infrared observations

Spaceborne precipitation radars provide the primary global observations of three-dimensional precipitation structure, yet their low-Earth-orbit sampling imposes substantial temporal gaps (Hou et al., 2014) that limit the ability to monitor rapidly evolving systems. Analysis of 151 days of observations reveals a mean radar revisit interval of 15.69 days, with more than 16% of regions experiencing gaps exceeding 20 days (Fig. 4b), highlighting a fundamental limitation in capturing the temporal evolution of precipitation systems.

By contrast, 4DPrecipNet leverages geostationary infrared observations to reconstruct 3D precipitation structure at high temporal frequency. For FY-4B/AGRI (Yang et al., 2017), retrievals are available at 15-minute intervals, reducing effective sampling gaps by nearly three orders of magnitude relative to radar observations. This enables continuous monitoring of precipitation-system evolution that is inaccessible to conventional radar-based approaches.

A representative case over the Indochina Peninsula (1 September 2025) demonstrates the ability of 4DPrecipNet to capture rapid intensification of a precipitation system between 07:00 and 08:00 UTC, with peak precipitation approaching 19 mm $h^{-1}$ (Fig. 4a). Independent radar observations from FY-3G/PMR and GPM-Core/DPR confirm the reconstructed vertical structure, with correlation coefficients of ~0.98 and ~0.99, respectively (Fig. 4c), indicating strong structural consistency across observing systems.

Further cross-platform evaluation using independent radar datasets outside the training period, conducted over a 7-day window from 1 to 7 September 2025, shows stable performance despite differences in sampling geometry and spatiotemporal coverage, with consistent correlation (~0.65) and detection skill (POD ~0.5) (Fig. 4d). Although the evaluation period is limited in duration, it encompasses a large number of independent dynamical samples, including 188 FY-3G overpasses and 109 GPM-Core overpasses, providing substantial statistical weight across diverse precipitation regimes. This agreement demonstrates robustness under heterogeneous observing conditions and confirms that the retrieved precipitation structures are physically consistent rather than sensor-specific artefacts. Together, these results establish that geostationary infrared observations can provide continuous, high-frequency constraints on the four-dimensional evolution of precipitation systems, complementing sparse radar measurements and enabling a new paradigm for global precipitation monitoring.

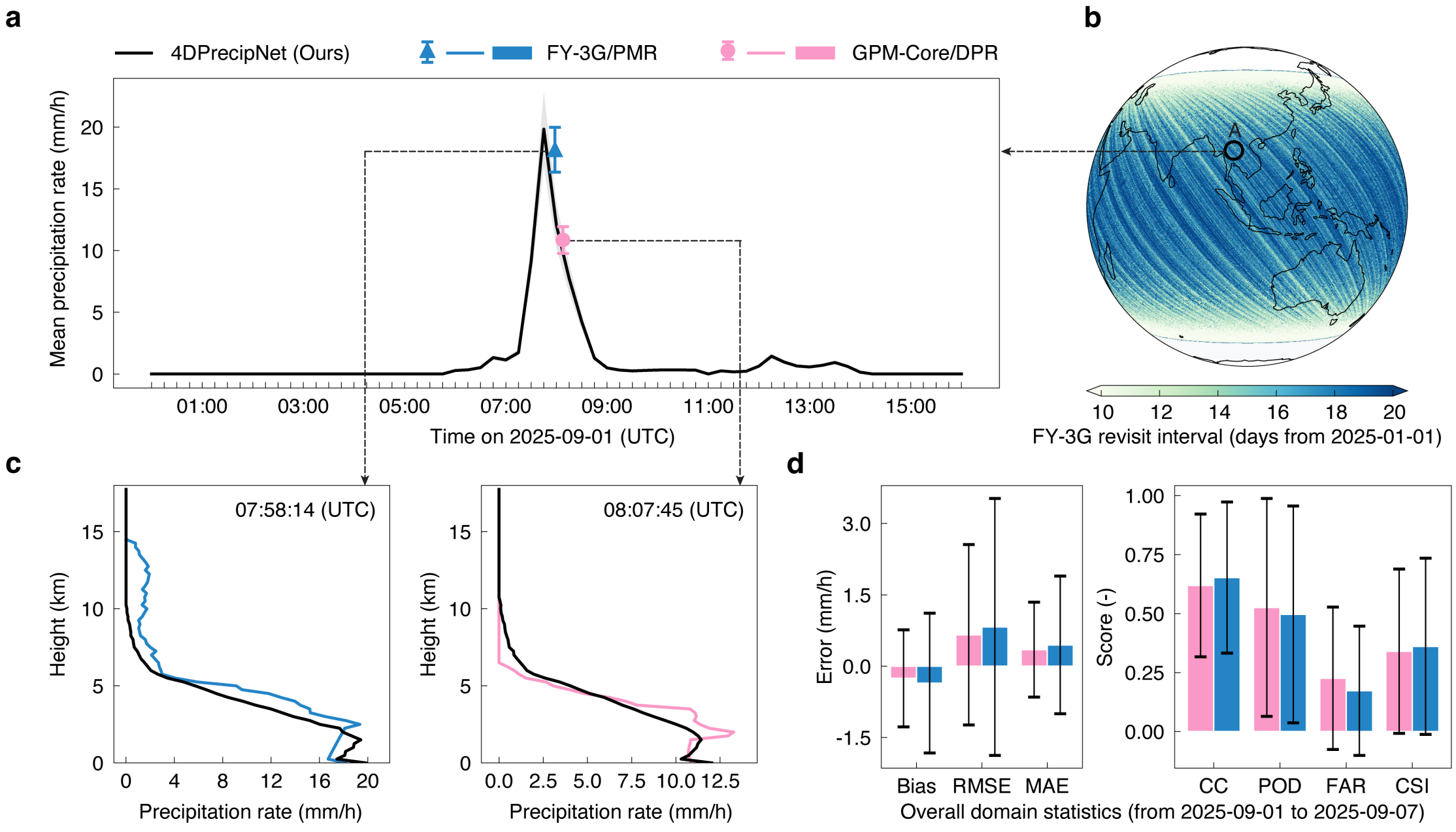

**Figure 4 | Continuous monitoring of precipitation-system evolution from geostationary infrared observations.**

**a,** Time series of mean precipitation rate for a precipitation system over the Indochina Peninsula (Point A: 100.72°E, 17.65°N) on 1 September 2025. 4DPrecipNet (black) captures rapid intensification after 07:00 UTC, with a peak approaching 19 mm h$^{-1}$, consistent with independent radar observations from FY-3G/PMR (blue) and GPM-Core/DPR (pink).

**b,** Spatial distribution of radar revisit intervals over 151 days, illustrating the temporal sparsity of low-Earth-orbit precipitation radar sampling, with extended gaps in many regions.

**c,** Comparison of vertical precipitation profiles at peak time. The 4DPrecipNet reconstruction (black) agrees closely with FY-3G/PMR (blue) and GPM-Core/DPR (pink), with correlation coefficients of ~0.98 and ~0.99, respectively, indicating strong cross-platform structural consistency.

**d,** Independent cross-platform evaluation (1–7 September 2025) against radar retrievals, showing consistent error and skill metrics across observing systems.

## 6. Discussion

The results presented here demonstrate that cloud-top infrared observations contain sufficient information to constrain the four-dimensional structure of precipitation, challenging the long-standing view that infrared measurements are intrinsically insensitive to sub-cloud hydrometeors. By integrating multi-channel infrared radiances with physically consistent atmospheric constraints, the proposed framework, 4DPrecipNet, reveals a previously underexploited information pathway linking cloud-top radiative signatures to precipitation processes beneath clouds. This establishes geostationary infrared observations as a viable source of 3D precipitation information, rather than solely a proxy for cloud-top properties.

A key implication of this work lies in the reinterpretation of infrared observability. Rather than being fundamentally limited, the apparent inability of infrared observations to resolve precipitation structure arises from the ill-posed nature of the inversion problem. By constraining the solution space through thermodynamic consistency and dynamical structure, 4DPrecipNet effectively stabilizes this inversion, enabling recovery of physically meaningful precipitation profiles (Raissi et al., 2019; Karniadakis et al., 2021). This perspective reframes infrared-based precipitation retrieval from an empirical estimation problem into a physically constrained inference problem, with broader implications for remote sensing of atmospheric processes.

The ability to reconstruct precipitation structure at geostationary scales provides a pathway to continuous monitoring of rapidly evolving weather systems. In contrast to low-Earth-orbit precipitation radars, which offer accurate but temporally sparse observations (Hou et al., 2014; Skofronick-Jackson et al., 2017), geostationary infrared sensors enable high-frequency sampling that captures the temporal evolution of convective systems. This capability is particularly relevant for extreme events, where rapid intensification and structural transitions occur on timescales that are poorly resolved by current observing systems (Houze, 2014). The framework presented here thus complements existing radar-based observations, bridging the gap between temporal continuity and vertical resolution.

Beyond weather-scale applications, continuous four-dimensional precipitation monitoring has important implications for understanding the global water and energy cycles (Loeb et al., 2024). Improved characterization of precipitation structure and evolution can enhance representation of moist processes in numerical weather prediction and climate models (Bi et al., 2023; Lam et al., 2023; Ben Bouallègue et al., 2024; Kochkov et al., 2024; Price et al., 2025; Radford et al., 2025), as well as support data assimilation of all-sky observations (Li et al., 2024; Schomburg et al., 2026). Furthermore, the ability to infer 3D precipitation from widely available infrared measurements opens opportunities for extending three-dimensional precipitation datasets to regions and time periods lacking radar coverage (Zhao et al., 2024; Katsanos et al., 2024).

Several limitations should be noted. The framework relies on physically informed constraints derived from reanalysis and radar observations (Hersbach et al., 2020), and its performance may be affected by biases in these reference datasets. While results demonstrate robustness across diverse meteorological regimes and observing conditions, uncertainties remain in weak-precipitation regimes, particularly for stratiform systems. This limitation arises from fundamental constraints in infrared observability: compared to deep convection, weak and stratiform precipitation is associated with limited vertical kinematic forcing and relatively uniform cloud-top thermal structure, such that cloud-top radiative signals provide only weak constraints on sub-cloud precipitation processes. The presence of extensive non-precipitating cirrus and multilayer cloud structures can further obscure these signals, making it difficult to reliably distinguish precipitating from non-precipitating clouds based solely on infrared observations. Additional uncertainties also arise in regions with complex surface emissivity or viewing geometry. Future work will focus on improving representation of light precipitation, extending validation to higher latitudes, and integrating the framework with data assimilation systems to fully exploit its potential for operational forecasting and climate applications.

## 7. Conclusion

We have shown that cloud-top infrared observations contain sufficient information to resolve the four-dimensional structure of precipitation, overturning the long-standing assumption that infrared measurements are intrinsically insensitive to processes beneath clouds. By integrating multi-channel infrared radiances with physically consistent constraints, the proposed framework recovers physically meaningful precipitation structures and their temporal evolution from geostationary platforms.

This capability bridges a fundamental gap in the current observing system: while spaceborne radars provide accurate but temporally sparse three-dimensional measurements, geostationary infrared sensors enable continuous, high-frequency three-dimensional monitoring. Together, these complementary strengths establish a new pathway for observing 4D precipitation systems across scales.

More broadly, the ability to infer 3D precipitation from widely available infrared observations opens new opportunities for improving representation of cloud–precipitation processes in weather and climate models, advancing all-sky data assimilation, and extending three-dimensional precipitation datasets to regions lacking radar coverage. These results suggest that geostationary infrared observations can serve as a foundation for continuous four-dimensional monitoring of the global water cycle.

# 8. Methods

## 8.1 Physically constrained Deep Learning Framework

### 8.1.1 The structure of 4DPrecipNet

4DPrecipNet (Figure 1c) is a physics-constrained encoder–decoder network (Ronneberger et al., 2015) that integrates multi-modal satellite observations and atmospheric priors to infer the three-dimensional structure of precipitation. The model takes as input nine-channel geostationary infrared brightness temperatures (3.7–13.3 μm) together with auxiliary geo-informational variables, including latitude, longitude, surface elevation, land-cover type, satellite viewing zenith angle, and local solar time. All radiometric inputs are first normalized using physically consistent ranges to account for sensor sensitivity and illumination effects.

The encoder (Thermodynamic Encoder) projects the 15-dimensional input feature space into a high-dimensional latent representation through successive convolutional blocks with down sampling. This process transforms top-of-atmosphere radiances into a thermodynamically meaningful moisture background state. Specifically, multi-scale convolutional layers expand the feature channels (from 15 to 1024) while reducing spatial resolution, enabling the network to encode large-scale moisture variability. The latent features are explicitly constrained to match reanalysis-derived precipitable water vapor (PWV) profiles (18 vertical layers) (Hersbach et al., 2020), providing a physically interpretable representation of atmospheric moisture.

At the bottleneck, a physically anchored global attention module is introduced to capture long-range dependencies and enforce global consistency. Self-attention layers operate on the compressed latent space (1024 × 32 × 8), enabling information exchange across spatial scales beyond local convolutional receptive fields. The bottleneck features are further regularized by PWV constraints, ensuring that the learned representations remain physically consistent with large-scale atmospheric states.

The decoder (Kinematics-driven Decoder) reconstructs the three-dimensional precipitation field through a sequence of upsampling and 3D precipitation blocks. Skip connections are used to fuse high-resolution cloud-top features from the encoder with dynamically evolving latent features. This design allows the model to learn the physical transformation from moisture availability to precipitation formation and vertical redistribution. The final outputs include precipitation intensity, vertical profiles (72 layers), and precipitation probability, generated through separate prediction heads.

To ensure physically consistent retrievals, a multi-dimensional physics-constrained loss function is applied, combining constraints on sensor sensitivity, geometric effects, precipitation intensity scaling, structural similarity, and probabilistic classification. This framework enables 4DPrecipNet to bridge cloud-top infrared observations with subsurface precipitation structures, yielding dynamically consistent and physically interpretable four-dimensional precipitation retrievals.

**8.1.2 Overview and Mathematical Formulation**

The objective of this study is to reconstruct the three-dimensional vertical structure of precipitation from two-dimensional multispectral geostationary satellite observations. Formally, the input field is defined as $X \in R^{H\times W\times C}$, and the target volumetric precipitation field as $Y \in R^{H\times W\times L}$. Here, $L$ =72 denotes vertical layers extending from the surface to 18 km altitude. The input feature dimension $C$ comprises nine infrared brightness temperature (TBB) channels and six static geo-information variables, including viewing zenith, latitude, longitude, elevation, landcover and solar time.

Physically, precipitation formation is governed by the coupling between thermodynamic moisture availability and dynamical vertical redistribution processes (Houze, 2014). We therefore adopt a decompositional hypothesis in which the three-dimensional precipitation field is generated through a two-step mechanism: first, large-scale PWV constrains the thermodynamic potential for condensation; second, dynamical structure redistributes this moisture vertically to produce the observed precipitation profile. Formally, this can be expressed as:

$$\boldsymbol{Y} = \mathcal{D}\left(\Phi_{\text{vap}}\big(\mathcal{E}(\mathbf{X})\big) + \Phi'_{\text{dyn}}(\boldsymbol{X})\right) \tag{1}$$

Here, $\mathcal{E}(\cdot)$ denotes a spectral encoder that maps raw satellite radiances into a latent feature space. $\Phi_{\text{vap}}(\cdot)$ represents a moisture-conditioned global bottleneck, realized through an auxiliary thermodynamic branch that explicitly encodes PWV states, while preserving the forward propagation of the main feature stream to the decoder. $\Phi'_{\text{dyn}}(\cdot)$ denotes the multi-scale dynamical residual component, thereby preserving structural gradients associated with convective uplift and vertical organization. Finally, $\mathcal{D}(\cdot)$ represents a three-dimensional recurrent decoder that integrates the thermodynamic background and dynamical residual features to reconstruct the vertical precipitation structure.

#### 8.1.3 The "Moisture-First" Constraint Mechanism

A key innovation of the framework lies in supervision of the latent bottleneck layer. Conventional autoencoder architectures allow latent representations to organize freely in feature space. In contrast, we explicitly constrain the deepest encoder layer—implemented as a dedicated thermodynamic branch—by requiring the encoded representation $\mathbf{Z} = \mathcal{E}(\mathbf{X})$ to solve an auxiliary PWV inversion task. This design forces the principal information bottleneck to concentrate on thermodynamic moisture potential, effectively quantifying how much condensable water is available for precipitation. Such a constraint is physically consistent with Clausius–Clapeyron scaling (Da Silva and Haerter, 2025), which governs the relationship between atmospheric moisture capacity and precipitation intensity. The extracted moisture representation, denoted as $\mathbf{Z}_{vap}$, is therefore defined as the thermodynamically constrained latent state derived from the encoder.

$$\mathbf{Z}_{vap} = \text{Bottleneck}(\mathcal{E}(\mathbf{X})) \tag{2}$$

To enforce physical consistency, a dedicated moisture-regression head projects the latent features to estimate the large-scale atmospheric moisture state. This estimation is supervised by a PWV loss term, $\mathcal{L}_{pwv}$, using ERA5 layer-integrated precipitable water as the reference target. This auxiliary constraint encourages the network to first diagnose the thermodynamic moisture condition before attempting to resolve complex vertical precipitation structure, thereby anchoring the retrieval process in physically consistent moisture states.

### 8.1.4 Dynamic Residual Injection and Convective Structure

Although atmospheric moisture sets an upper bound on precipitation potential, the realization of rainfall—particularly deep convective precipitation—is dynamically driven by vertical motions, including updrafts and downdrafts. These dynamical processes are manifested in satellite imagery as cloud-texture variability, cloud-top roughness and thermal gradients.

To recover such information, we introduce a residual dynamical pathway, $\Phi'_{\text{dyn}}(\boldsymbol{X})$, implemented through multi-scale skip connections. This component captures high-frequency structural features associated with convective organisation, complementing the thermodynamic constraint branch and enabling reconstruction of vertically redistributed precipitation structure.

$$\Phi'_{\text{dyn}}(\boldsymbol{X}) = \sum_{k} \boldsymbol{S}^{(\boldsymbol{k})} \tag{3}$$

Here, $\boldsymbol{S}^{(\boldsymbol{k})}$ denotes the feature map extracted from the k-th downsampling stage of the encoder. These residual features propagate uncompressed spatial information—such as cloud texture and overshooting tops—directly to the decoder, thereby reinforcing dynamically active structures.

During decoding, the model integrates the thermodynamic background representation, $\boldsymbol{Z}_{\text{vap}}$ (encoding moisture potential), with the set of dynamical residual features $\{\boldsymbol{S}^{(\boldsymbol{k})}\}_{k=1}^{K}$ (encoding convective forcing), to synthesise the three-dimensional precipitation field $\widehat{\boldsymbol{Y}}$.

$$\widehat{\boldsymbol{Y}} = \text{Decoder}(\boldsymbol{Z}_{\text{vap}} \oplus \{\boldsymbol{S}^{(\boldsymbol{k})}\}_{k=1}^{K}) \tag{4}$$

Here, $\widehat{\boldsymbol{Y}}$ denotes the reconstructed three-dimensional precipitation field, with $H \times W \times L$ dimensions and $L$ = 72 vertical layers. $\boldsymbol{Z}_{\text{vap}}$ denotes the thermodynamically constrained latent representation derived from the bottleneck layer (Eq. 2). The set $\{\boldsymbol{S}^{(\boldsymbol{k})}\}$ from $k = 1$ to $K$ denotes the multi-scale dynamical residual features extracted from the encoder stages (Eq. 3), and $\oplus$ denotes feature-wise fusion through skip connections and channel concatenation.

## 8.2 Network Architecture Implementation

The implementation adopts a hierarchical symmetric encoder–decoder architecture tailored for three-dimensional volumetric generation.

(1) **Thermodynamic Encoder.** Static geographical variables are first mapped into a physical potential embedding. The encoder then operates as a four-stage convolutional downsampling network (channels: 128 → 256 → 512 → 1024) using GeLU activations and batch normalization to extract multi-scale representations (He et al., 2016). The large effective receptive field enables capture of mesoscale organization, including squall lines and spiral rainbands.

(2) **Physically Anchored Global Bottleneck.** At the lowest spatial resolution (one-eighth of the input size), a global self-attention module integrates large-scale features (Vaswani et al., 2017) under strict supervision from the auxiliary PWV loss. This bottleneck performs global feature aggregation while enforcing thermodynamic consistency. Dedicated three-dimensional precipitation blocks are incorporated within both encoder and decoder stages to enhance spatiotemporal feature extraction.

(3) **Kinematics-Driven Decoder.** The decoder transforms two-dimensional latent representations into three-dimensional vertical profiles. Because the output dimension (H × W × 72) requires explicit physical mapping, a dedicated regression head projects feature channels from 128 to 72, corresponding to predefined atmospheric height layers.

(4) **Output Heads.** The architecture includes a value head for normalized rain-rate regression, a probability head for precipitation masking, and an auxiliary PWV head for physical regularization.

(5) **Efficiency and Scalability.** The model contains 73,639,474 parameters and remains computationally lightweight, with peak training memory usage of approximately 23 GB. End-to-end training can be performed on a single NVIDIA RTX 4090 GPU. The framework further supports FD three-dimensional inference on FY-4B AGRI data without tiling, demonstrating feasibility for operational deployment. Increased model capacity and multi-GPU high-memory training are expected to further enhance performance.

## 8.3 Loss Landscape and Optimization

Model training is guided by a multi-objective loss function, $\mathcal{L}_{total}$, which is decomposed into two principal components: a sensor-physics consistency loss that accounts for intrinsic limitations of satellite infrared observations, and a precipitation reconstruction loss that constrains the statistical fidelity of the retrieved rainfall field. In addition, a PWV loss term is incorporated as an explicit physical regularization to enforce thermodynamic consistency within the latent representation.

$$\mathcal{L}_{total} = \underbrace{\lambda_{sens}\mathcal{L}_{sens} + \lambda_{geo}\mathcal{L}_{geo}}_{\text{Sensor Physics}} + \underbrace{\lambda_{scale}\mathcal{L}_{scale} + \lambda_{struct}\mathcal{L}_{struct} + \lambda_{prob}\mathcal{L}_{prob}}_{\text{Precipitation Field}} + \lambda_{pwv}\mathcal{L}_{pwv} \tag{5}$$

1. Observation sensor-physics constraints

• Light-rain suppression (sensitivity) loss, $\mathcal{L}_{sens}$ : introduced to prevent the model from generating spurious precipitation signals near the observational noise floor of infrared measurements.

$$\mathcal{L}_{sens} = \frac{1}{N_{noise}} \sum_{i \in \Omega_{noise}} (\hat{y}_i)^2 \, , \, \Omega_{noise} = \{ i \mid y_i < \epsilon \wedge \hat{y}_i > \tau_{small} \} \tag{6}$$

Here, $N_{noise}$ denotes the number of pixels within the noise subset $\Omega_{noise}$. The parameter $\epsilon$ represents the minimum threshold for true rainfall, set to the operational standard of 0.1 mm/h, and $\tau_{small}$ defines the suppression threshold applied to constrain light-precipitation estimates.

• Geolocation-uncertainty loss, $\mathcal{L}_{geo}$ : formulated to account for physical uncertainties associated with satellite viewing zenith angle ($\theta_{zen}$) and latitude ($\phi_{lat}$).

$$W_{geo} = \exp\left( \text{clip}\left( \frac{\alpha}{\cos(\theta_{zen}) + \epsilon} + \frac{\beta}{\cos(\phi_{lat}) + \epsilon}, 0, C_{max} \right) \right) \tag{7}$$

Here, $\alpha$ and $\beta$ control the distortion curvature associated with satellite zenith angle and latitude, respectively, and $C_{max}$ denotes the truncation constant.

$$\mathcal{L}_{geo} = \frac{1}{N} \sum \left( \hat{p} \cdot W_{geo} \right) + \frac{1}{N} \sum \left( \hat{y} \cdot W_{geo} \right) \tag{8}$$

Here, $N$ denotes the total number of samples, $\hat{p}$ represents the predicted precipitation probability, and $\hat{y}$ denotes the predicted rain-rate value. This loss term penalises erroneous predictions in regions of high uncertainty.

• Scale-aware intensity loss, $\mathcal{L}_{scale}$: a dynamically weighted L2 loss designed to emphasize high-intensity precipitation events.

$$\mathcal{L}_{scale} = \frac{1}{N}\sum w_i \cdot ||\hat{y}_i - y_i||_2^2,\ w_i = \max\left(1, \lambda_{ext} \cdot \left(\frac{y_i}{\tau_{ext}}\right)^{\beta}\right) \tag{9}$$

Here, $y_i$ and $\hat{y}_i$ denote the true and predicted values, respectively; $w_i$ represents the sample-specific weight; $\lambda_{ext}$ controls the amplification of extreme events; $\tau_{ext}$ defines the threshold for extreme-event classification; and $\beta$ denotes the power exponent.

• Structural composite loss, $\mathcal{L}_{struct}$: integrates structural similarity (SSIM) (Wang et al., 2004), correlation-based constraints, and gradient consistency terms to preserve spatial organization and fine-scale precipitation features.

$$\mathcal{L}_{struct} = \lambda_{SSIM}\mathcal{L}_{SSIM} + \lambda_{CC}\left(1 - \rho(\hat{Y}, Y)\right) + \lambda_{Grad}||\nabla\hat{Y} - \nabla Y||_1 \tag{10}$$

Here, $\lambda_{SSIM}$, $\lambda_{CC}$, and $\lambda_{Grad}$ denote the weighting coefficients for the structural similarity index (SSIM), the Pearson correlation coefficient ($\rho$), and the gradient term ($\nabla$), respectively.

• Precipitation probability classification loss, $\mathcal{L}_{prob}$: implemented as a binary cross-entropy loss to supervise rainfall occurrence prediction.

$$\mathcal{L}_{prob} = -\frac{1}{N}\sum[y_{bin}\log(\hat{p}) + (1 - y_{bin})\log(1 - \hat{p})] \tag{11}$$

Here, $y_{bin}$ denotes the binarised ground-truth label (1 indicating rainfall occurrence and 0 indicating no rainfall), and $\hat{p}$ represents the predicted precipitation probability.

## 8.4 Impact of Thermodynamic Constraints

To quantify the contribution of the moisture-constraint design, we conducted a systematic ablation study (Supplementary Fig. 1), training 11 model variants with different thermodynamic loss weights ($\lambda_{pwv}$ ranging from 0 to 10). The results reveal a clear optimal regime around $\lambda_{pwv}$ = 5, characterised by a distinct performance inflection point.

**Bias correction (thermodynamic gating effect).** Without thermodynamic constraints ($\lambda$ = 0), the model exhibits a pronounced positive bias (+0.040), frequently misclassifying cold cloud tops as precipitation. Introducing the optimal constraint ( $\lambda$ = 5) reduces this systematic error by approximately 20% (Bias ~ -0.0007), effectively acting as a thermodynamic gate that suppresses false detections in moisture-deficient environments.

**Skill enhancement.** The POD increases from ~ 0.60 in the unconstrained model to ~ 0.64 under optimal constraint, indicating improved sensitivity to precipitation events when thermodynamic consistency is enforced.

**Risk of over-constraint.** When the constraint weight becomes excessive ($\lambda$ = 10), performance degrades, with RMSE increasing to 0.44. This suggests that overemphasis on large-scale moisture consistency may suppress fine-scale vertical structure reconstruction.

Based on a weighted evaluation criterion (80% vertical-structure accuracy and 20% surface rainfall accuracy), $\lambda_{pwv}$ = 5 is adopted as the default configuration.

# 9 Data Acquisition and Processing

## 9.1 Multi-Source Data Fusion and Tensor Construction

To mitigate parallax effects and displacement errors in convective systems, we developed a robust spatiotemporal matching framework based on a KD-tree nearest-neighbour algorithm. The training dataset comprises three principal components (in addition to static geographical information):

(1) **Geostationary satellite inputs.**

Approximately 1.5 years of FY-4B/AGRI Level-2 brightness temperature (TBB) observations (Yang et al., 2017) from 5 March 2024 to 31 August 2025. Nine infrared channels are used as input features and combined with static geographical variables to form the model input tensor.

(2) **Active radar reference.**

Precipitation retrievals from FY-3G/PMR serve as the primary training reference (Zhang et al., 2019). Strict collocation criteria are applied, with a spatial radius of 0.1° and a temporal window of 600 seconds.

(3) **Thermodynamic constraint data.**

ERA5 pressure-level data obtained from the DestinE Earth Data Hub (Hersbach et al., 2020) are incorporated, providing 19 vertical layers of specific humidity as thermodynamic prior information.

## 9.2 Physics Feature Engineering: Layer-wise PWV Integration

Rather than relying on a single-column PWV value, we compute vertically resolved PWV slices to guide the latent bottleneck (thermodynamic branch). Using specific humidity ($q$) from 19 ERA5 pressure levels (1000, 925, 850, 700, 600, 500, 400, 300, 250, 200, 150, 100, 70, 50, 30, 20, 10, 5 and 1 hPa), we integrate moisture content across 18 atmospheric layers.

For each layer bounded by pressures $P_i$ and $P_{i+1}$, the corresponding PWV contribution is computed via hydrostatic integration.

$$PWV_i = \frac{1}{g \cdot \rho_w} \int_{P_i}^{P_{i+1}} q(p)\ dp \sim \frac{\overline{q_i} \, \Delta P_i}{g \cdot \rho_w} \tag{12}$$

Here, $g$ denotes gravitational acceleration and $\rho_w$ represents the density of water. This formulation yields an 18-channel physically derived moisture tensor, explicitly quantifying the latent moisture potential available within each atmospheric layer.

The final training dataset comprises 14,036 high-quality matched observations (~ 663 GB). For cross-sensor evaluation, an independent test dataset was constructed using all successfully matched observations between 1 and 7 September 2025 from FY-3G/PMR (188 overpasses) and GPM-Core/DPR (109 overpasses).

## 9.3 Experimental Design and Benchmarking

To ensure rigorous independent evaluation, we adopted the following experimental protocols.

(1) **Dataset partitioning.**

The dataset was randomly divided into training and validation subsets at a 9:1 ratio. To mitigate the rarity of extreme precipitation events, a weighted sampler was employed during training to prioritise high-intensity rainfall samples.

(2) **Independent validation strategies.**

• **Vertical structure validation (GPM-Core/DPR).**

Unseen FY-4B observations from the training period were collocated with corresponding GPM-DPR overpasses (1,260 randomly sampled orbits). The radar profiles were interpolated to 72 vertical layers to ensure consistent dimensional comparison.

• **Cross-sensor independent validation (FY-3G/PMR and GPM-Core/DPR).**

An independent test dataset was constructed using all successfully collocated observations between 1 and 7 September 2025 from FY-3G/PMR (188 overpasses) and GPM-Core/DPR (109 overpasses). This dataset was used to evaluate cross-sensor consistency and temporal generalization capability.

• **Surface rainfall validation (IMERG-Final).**

The widely used IMERG-Final product served as the reference baseline. Evaluation was conducted using 105 hours of unseen FY-4B observations (randomly sampled) corresponding to the validation period. To ensure fair comparison, all products—including satellite retrievals and IMERG-Final—were resampled to a common temporal resolution (1 hour) and spatial grid (0.1°).

(3) **Ablation study configuration.**

The moisture-constraint ablation experiments (Supplementary Fig. 1) were performed using models trained for one epoch. Validation was conducted on a reduced subset (20 surface snapshots and 180 vertical-profile orbits) to enable efficient comparative analysis.

## 9.4 Comparison Product Specifications

To ensure spatial and temporal consistency across datasets, all products were strictly aligned to the IMERG-Final 0.1° reference grid. Collocation was performed using a KD-tree nearest-neighbor search (matching radius < 0.1°), and only temporally overlapping observations were retained to minimize sampling bias. Product specifications are summarized below.

• **IMERG-Final V07B (reference baseline).**

A multi-satellite integrated precipitation product with an original spatial resolution of 0.1° × 0.1° and 30-minute temporal resolution. The "Final Run" version incorporates passive microwave observations and retrospective gauge-based bias correction using GPCC monthly analyses, representing one of the most accurate and heavily corrected global precipitation benchmarks currently available. In this study, consecutive 30-minute estimates were averaged to derive hourly mean precipitation fields consistent with the temporal sampling of the geostationary products (Huffman et al., 2020).

• **FY-4B/AGRI QPE (operational product 1).**

An operational quantitative precipitation estimate released by the National Satellite Meteorological Center (NSMC). The original spatial resolution is 4 km with hourly output. The retrieval algorithm is based on machine-learning approaches (random forest and multilayer perceptron hybrid), trained and calibrated using sparse surface rain-gauge observations (Huang et al., 2024).

• **FY-4B/AGRI PECA (operational product 2).**

A cloud-property-based precipitation estimation product with an original spatial resolution of 4 km and 15-minute temporal resolution (aggregated to hourly in this study). The retrieval is grounded in physical cloud-top property inversion methods (Zhu and Ma, 2022).

• **PERSIANN-CCS (operational product 3).**

An infrared-based precipitation estimation and cloud classification system built on artificial neural networks. The original resolution is 0.04° × 0.04° with hourly output. The algorithm relies on cloud-image segmentation and temperature–rainfall mapping relationships (Hong et al., 2004).

## Data availability

The FY-4B AGRI L2 TBB data and FY-3G/PMR data are publicly available from the National Satellite Meteorological Center (NSMC) (http://www.nsmc.org.cn). GPM-Core/DPR and IMERG-Final data are available from the NASA PMM data archive (https://gpm.nasa.gov/data). The ERA5 Pressure Level data can be accessed via the DestinE EarthDataHub. The processed matchups and 3D volumetric labels generated during this study are available from the corresponding author upon reasonable request.

## Code availability

The code for model inference, testing, and visualization is available at https://github.com/PKU-Precip/4DPrecipNet.git. To facilitate reproduction and application, pre-trained model weights and example data processing scripts are provided in the repository.

## Acknowledgments

We acknowledge the National Satellite Meteorological Center (NSMC) for providing the FY-4B AGRI and FY-3G/PMR datasets; the National Aeronautics and Space Administration (NASA) for the GPM-Core/DPR and IMERG-Final products; the Center for Hydrometeorology and Remote Sensing (CHRS) at the University of California, Irvine, for the PERSIANN-CCS precipitation dataset; and the European Centre for Medium-Range Weather Forecasts (ECMWF) for the ERA5 reanalysis data. This work was supported by the National Natural Science Foundation of China (Grant No. 42371335); the National Key R&D Program of China (Grant No. 2023YFB3905900); Sichuan Science and Technology Program (No.2025YFNH0001, No.2025YFNH0006); the National Natural Science Foundation of China (Grant No. 41901343); the China Postdoctoral Science Foundation (No. 2018M630037, and 2019T120021); the Fengyun Application Pioneering Project (FY-APP).

## Author Contributions

T.X. and Z.M. conceived the study. T.X. developed the deep learning framework and performed the analyses. A.M. provided scientific guidance during the research. Y.He. further upgraded and optimized the framework. X.L. provided data support. N. L., C.Z., K.H., J.X., T.W. and D.W. provided suggestions on revising the manuscript. B.Z. provided conceptual suggestions on the framework design. W.Z. and H.C. assisted with data preparation. D.W. and Z.M. provided computational resources support. Y.Ho. critically reviewed and revised the manuscript. Z.M. and T.X. drafted and revised the manuscript. Z.M. supervised the study and takes responsibility for the integrity of the work as a whole. All authors contributed to the scientific interpretation of the results and the preparation of the manuscript.

## Competing Interests

The authors declare no competing interests.

## Extended Data Figures

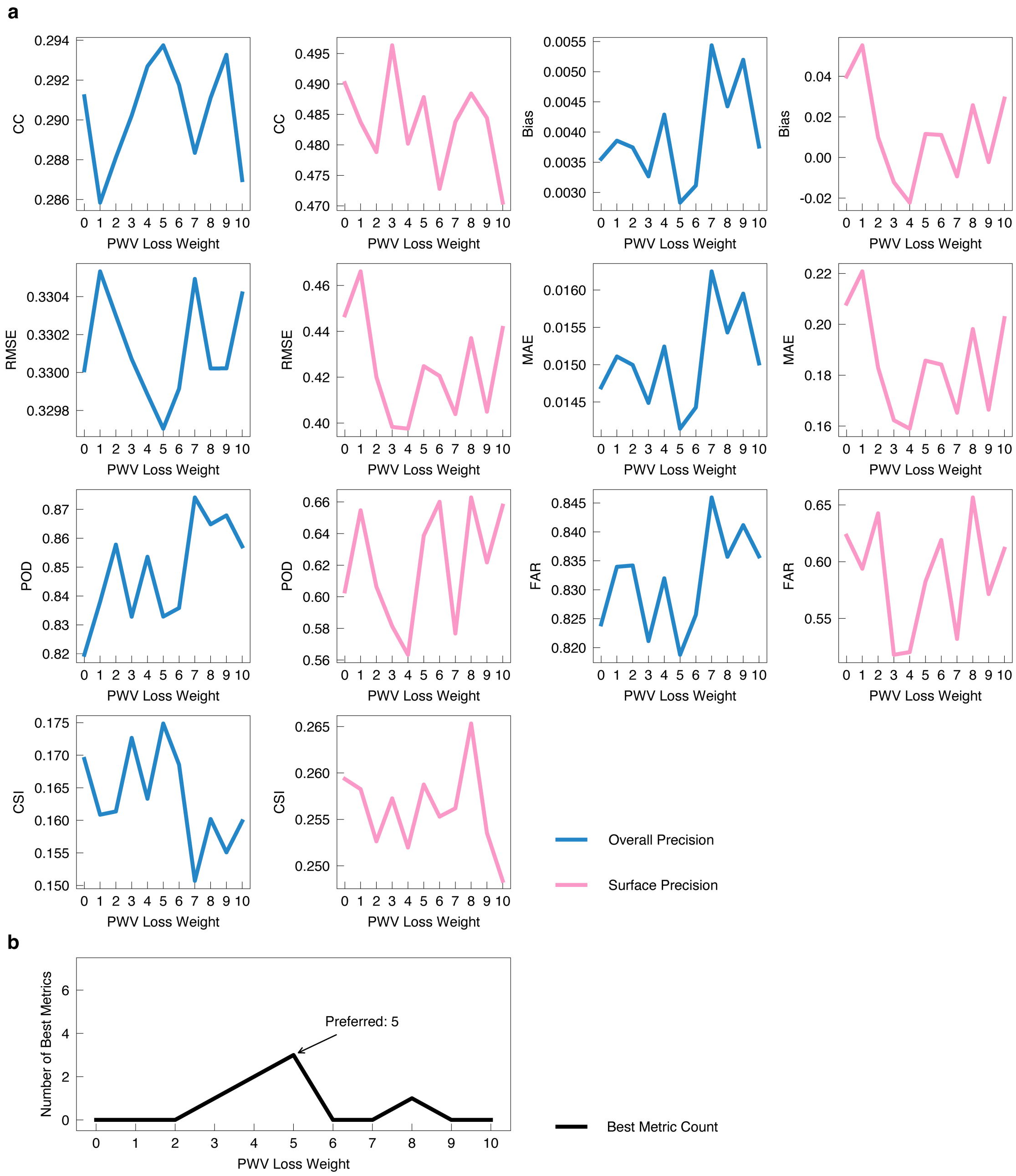

**Supplementary Figure 1 | Ablation analysis of thermodynamic (PWV) constraints.**

(a) Sensitivity analysis of performance metrics, showing variations in CC, bias, RMSE, MAE, POD, FAR and CSI as a function of the PWV loss weight ($\lambda_{pwv}$ ranging from 0 to 10). Results are based on models trained for one epoch and evaluated on a validation subset comprising 180 vertical-profile orbits and 20 surface snapshots.

(b) Distribution of composite evaluation scores under a weighted assessment scheme. The bar plot and integrated score curve (black line) indicate that $\lambda = 5$ yields the most stable overall performance (highest composite score), representing an optimal balance between physical constraint and data-driven representation learning.

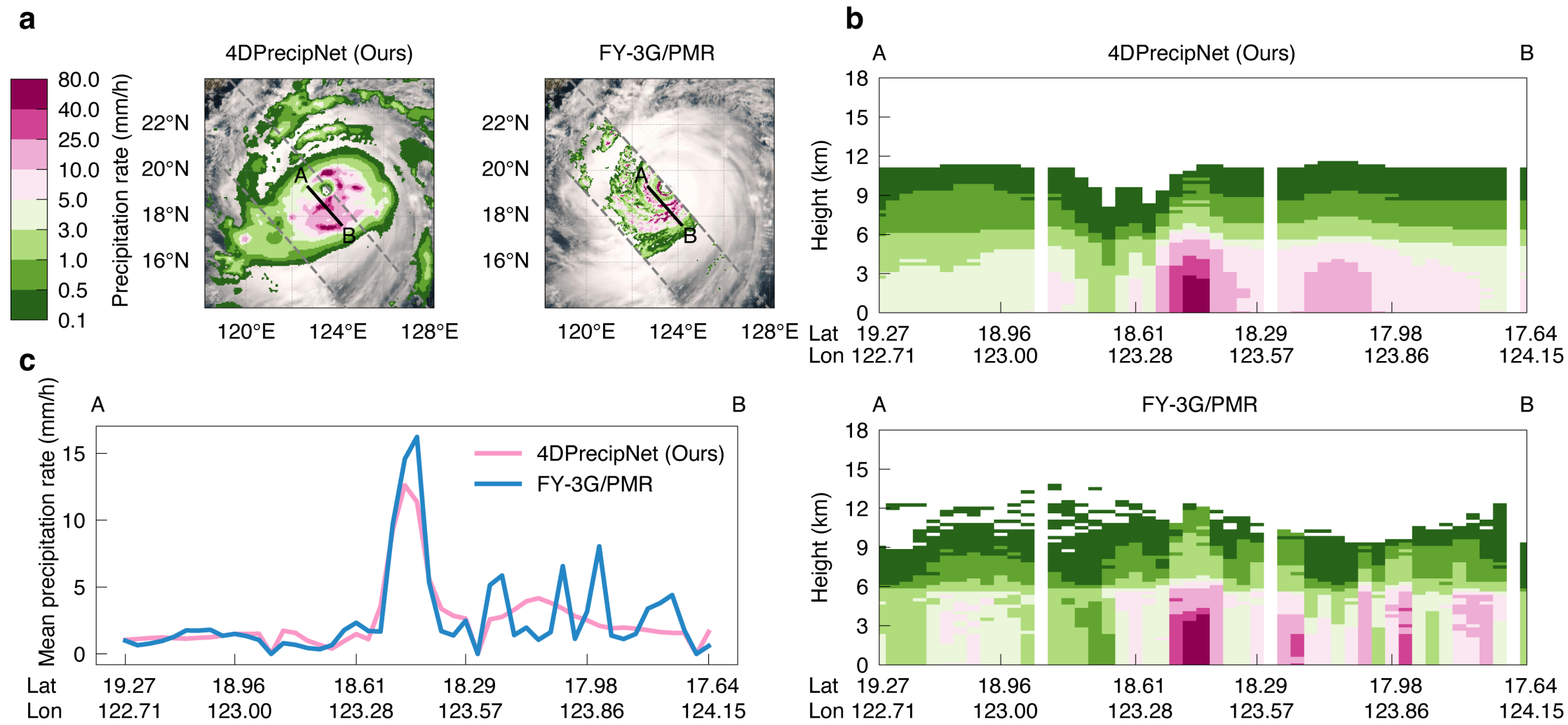

**Supplementary Figure 2 | Three-dimensional structural analysis of the asymmetric Typhoon Ragasa.** Extended Data Fig. 2 presents retrieval results for Typhoon Ragasa, characterized by a loosely organized and asymmetric convective structure. Despite the absence of a well-defined eyewall configuration typical of mature tropical cyclones, 4DPrecipNet resolves multiple embedded convective cores and reproduces their distinct vertical extent. Notably, three precipitation-free zones identified in the retrieved vertical cross-section are consistent with radar-observed downdraft regions (Extended Data Fig. 2b). This agreement suggests that the model captures physically meaningful dynamical features associated with subsidence between convective cells, rather than merely smoothing spatial patterns. This case demonstrates that the framework does not rely on memorization of canonical storm morphology (for example, symmetric eyewall structures), but instead infers precipitation organization through dynamically informed structural reconstruction, enabling generalization to atypical convective systems.

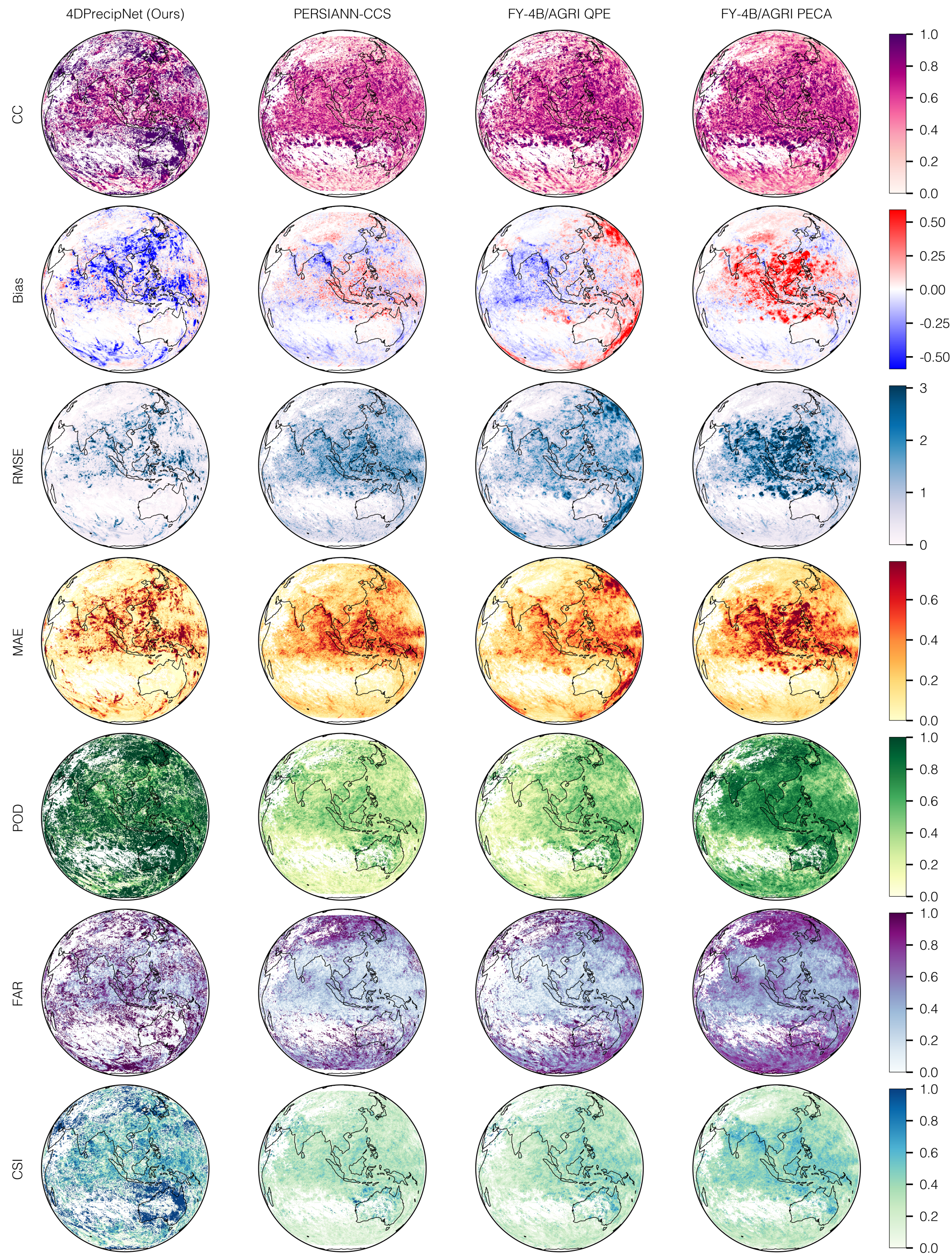

**Supplementary Figure 3 | Spatial distribution of full-disk surface precipitation retrieval performance against IMERG-Final.**

The figure compares the proposed 4DPrecipNet model (first column) with three baseline products (PERSIANN-CCS, FY-4B/AGRI QPE and FY-4B/AGRI PECA) over 18 validation months (March 2024 to August 2025) across the full disk domain (30°E–175°E, 75°S–75°N). Pixel-wise statistical metrics are shown.

(1) **Spatial consistency (CC and RMSE).** 4DPrecipNet exhibits strong spatial coherence, with correlation coefficients (CC) exceeding 0.6 over the Qinghai–Tibet Plateau and adjacent oceanic regions, while maintaining low RMSE (< 0.8 mm/h). In contrast, the FY-4B/AGRI QPE product shows elevated errors over the Bay of Bengal and the South Pacific (RMSE > 2.5 mm/h).

(2) **Bias control.** 4DPrecipNet maintains spatially homogeneous bias across the domain, with fluctuations within ±0.05 mm/h. By comparison, PERSIANN-CCS exhibits patch-like bias patterns, likely reflecting limitations of its cloud-classification-based retrieval approach.

(3) **Topographic robustness (CSI).** In complex terrain such as the Qinghai–Tibet Plateau and the Himalayan region, baseline products (FY-4B/AGRI PECA and PERSIANN-CCS) show reduced skill (CSI < 0.1). In contrast, 4DPrecipNet retains CSI values of ~ 0.35, demonstrating enhanced robustness in extreme topographic environments.

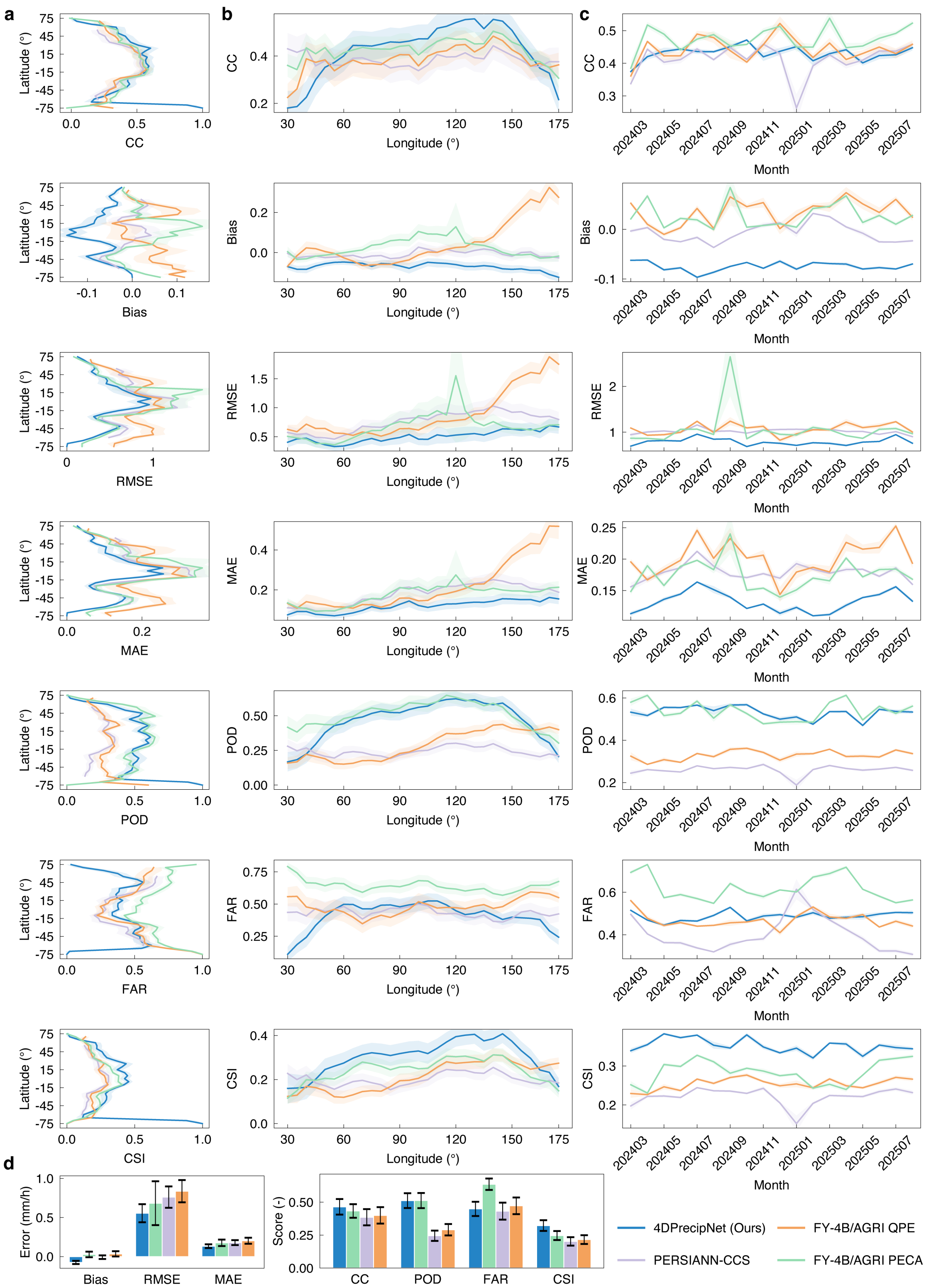

**Supplementary Figure 4 | Full-disk spatiotemporal consistency of surface precipitation retrieval skill against IMERG-Final.**

(a) **Latitudinal performance distribution.** Statistics aggregated along latitude bands indicate that 4DPrecipNet maintains stable performance across tropical and subtropical regions (30°S–30°N), with correlation coefficients (CC) exceeding 0.5 and critical success index (CSI) values of approximately 0.35. In contrast, infrared-only baseline products (PERSIANN-CCS and FY-4B/AGRI PECA) exhibit degradation in subtropical zones. Owing to its physical constraints, 4DPrecipNet retains effective detection skill (CSI > 0.2) even at 50°N.

(b) **Longitudinal performance distribution.** Within the observational domain (60°E–150°E), performance curves remain comparatively smooth across longitude, indicating robustness to viewing-angle variations and reduced sensitivity to edge-of-disk degradation effects commonly observed in conventional retrieval approaches.

(c) **Temporal evolution.** Eighteen-month time series analysis demonstrates strong seasonal stability, with CSI standard deviation of approximately 0.02. Unlike PERSIANN-CCS, which exhibits noticeable winter degradation, 4DPrecipNet maintains consistent performance throughout all seasons, reflecting improved generalisation across precipitation regimes.

(d) **Global aggregated comparison.** Global summary metrics indicate that 4DPrecipNet achieves the lowest overall error (RMSE ~ 0.55 mm/h) while improving detection capability by more than 30% relative to operational baselines (CSI 0.32 versus 0.24). These results suggest that the framework establishes a new performance benchmark for geostationary precipitation retrieval.